\documentclass[conference]{IEEEtran}
\usepackage[latin9]{inputenc}
\usepackage{amsmath}
\usepackage{amssymb}
\usepackage{graphicx}

\makeatletter

\providecommand{\tabularnewline}{\\}

\IEEEoverridecommandlockouts
\usepackage{cite}
\usepackage{amsfonts}\usepackage{algorithmic}
\usepackage{textcomp}
\usepackage{xcolor}
\def\BibTeX{{\rm B\kern-.05em{\sc i\kern-.025em b}\kern-.08em
    T\kern-.1667em\lower.7ex\hbox{E}\kern-.125emX}}

\usepackage{amsmath,amssymb}
\DeclareMathOperator{\E}{\mathbb{E}}

\makeatother

\begin{document}
\title{Tail of Distribution GAN (TailGAN): Generative- Adversarial-Network-Based
Boundary Formation}
\author{\IEEEauthorblockN{Nikolaos Dionelis, Mehrdad Yaghoobi, Sotirios A. Tsaftaris}
\IEEEauthorblockA{\textit{The University of Edinburgh} \\
 \textit{Edinburgh}, UK \\
 Contact email: nikolaos.dionelis@ed.ac.uk} }
\maketitle
\begin{abstract}
Generative Adversarial Networks (GAN) are a powerful methodology and
can be used for unsupervised anomaly detection, where current techniques
have limitations such as the accurate detection of anomalies near
the tail of a distribution. GANs generally do not guarantee the existence
of a probability density and are susceptible to mode collapse, while
few GANs use likelihood to reduce mode collapse. In this paper, we
create a GAN-based tail formation model for anomaly detection, the
Tail of distribution GAN (TailGAN), to generate samples on the tail
of the data distribution and detect anomalies near the support boundary.
Using TailGAN, we leverage GANs for anomaly detection and use maximum
entropy regularization. Using GANs that learn the probability of the
underlying distribution has advantages in improving the anomaly detection
methodology by allowing us to devise a generator for boundary samples,
and use this model to characterize anomalies. TailGAN addresses supports
with disjoint components and achieves competitive performance on images.
We evaluate TailGAN for identifying Out-of-Distribution (OoD) data
and its performance evaluated on MNIST, CIFAR-10, Baggage X-Ray, and
OoD data shows competitiveness compared to methods from the literature.
\end{abstract}


\section{Introduction}

Generative Adversarial Networks (GAN) can capture complex data with
many applications in computer vision and have achieved state-of-the-art
image synthesis performance. Recently, GANs have been used for anomaly
detection which is critical in security, e.g. contraband detection.
GANs succeed convergence in distribution metrics, but present limitations
as they suffer from mode collapse, learn distributions of low support,
and do not guarantee the existence of a probability density making
generalization with likelihood impossible \cite{AdamOptimizer-2-1-1}.
Unsupervised anomaly detection is examined since anomalies are not
known in advance \cite{Deecke}. The normal class is learned and abnormal
data are detected by deviating from this model.

Many anomaly detection methods perform well on low-dimensional problems;
however, there is a lack of effective methods for high-dimensional
spaces, e.g. images. Important tasks are reducing false negatives
and false alarms, providing boundaries for inference of within-distribution
and Out-of-Distribution (OoD), and detecting anomalies near low probability
regions. Most papers use the leave-one-out evaluation methodology,
$(K + 1)$ classes, $K$ classes for normality, and the leave-out
class for anomaly, chosen arbitrarily. This evaluation does not use
the complement of the support of the distribution, and real-world
anomalies are not confined to a finite set.

In this paper, we create a GAN-based model, the Tail of distribution
GAN (TailGAN), to generate samples on the low probability regions
of the normal data distribution and detect anomalies close to the
support boundary. Using TailGAN, we leverage GANs for OoD sample detection
and perform sample generation on the tail using a cost function that
forces the samples to lie on the boundary while optimizing an entropy-regularized
loss to stabilize training. The authors of this paper have recently
proposed an invertible-residual-network-based generator, the Boundary
of Distribution Support Generator (BDSG) \cite{AdamOptimizer-1},
for anomaly detection using the IResNet and ResFlow models {[}6,~7{]}.
In this paper, we include adversarial training and GANs to better
generate samples on the low probability regions of the data distribution
and detect anomalies near the support boundary. Our contribution is
the creation of a GAN-based boundary formation model and we use GANs,
such as Prescribed GAN (PresGAN) and FlowGAN {[}9,~10{]}, that learn
the probability of the underlying distribution and generate samples
with high likelihoods. TailGAN improves the detection of anomalies
and GANs that learn the probability density of the underlying distribution
improve the anomaly detection methodology, allowing us to create a
generator for boundary samples and use this to characterize anomalies.

\section{Related Work on Anomaly Detection}

\subsection{GANs Using a Reconstruction-Based Anomaly Score}

AnoGAN performs unsupervised learning to detect anomalies by learning
the manifold of normal anatomical variability. During inference, it
scores image patches indicating their fit into the learned distribution
{[}11,~12{]}. In contrast to AnoGAN, Efficient GAN-Based Anomaly
Detection (EGBAD) jointly learns an encoder and a generator to eliminate
the procedure of computing the latent representation of a test sample
\cite{BiGAN}. Its anomaly score combines discriminator and reconstruction
losses. GANomaly learns the generation of the data and the inference
of the latent space, $\textbf{z}$. It uses an encoder-decoder-encoder
in the generator and minimizing the distance between the vectors in
$\textbf{z}$ aids in learning the data distribution \cite{GANomaly}.

\subsection{GANs Performing Sample Generation on the Boundary}

The GAN loss learns the mass of the distribution but to perform anomaly
detection, we look at the boundary. The GAN optimization problem is
given by $\text{argmin}_{\pmb{\theta}_g} \text{dist}( p_{\textbf{x}}(\textbf{x}), p_g(\textbf{x}) )$,
where the distance metric, $\text{dist}(.,.)$, takes the specific
form $\text{argmax}_{\pmb{\theta}_d} \, f( D, p_{\textbf{x}}(\textbf{x}), p_g(\textbf{x}) )$.
For example, $\text{dist}(\textbf{x}, \textbf{y}) = ||\textbf{x} - \textbf{y} ||_{\infty} = \text{max}_i \, | x_i - y_i |$
and $\textbf{x},\textbf{y} \in \mathbb{R}^d$. The GAN loss is \vspace{-1.8pt}
\begin{align}
\begin{split}
\text{min}_{G} & \, \text{max}_{D} \E_{\textbf{x}} [\text{log}(D(\textbf{x}))] + \E_{\textbf{z}} [\text{log}(1 - D(G(\textbf{z})))]
\end{split}
\end{align}where $\textbf{z} \sim p_{\textbf{z}}(\textbf{z}) = N(\textbf{0}, \textbf{I})$, $\textbf{x} \sim p_{\textbf{x}}(\textbf{x})$, and $G(\textbf{z}) \sim p_g(\textbf{x})$.
To generate samples on the distribution tail, MinLGAN uses minimum
likelihood while FenceGAN changes (1). In contrast to the traditional
GAN {[}17,~18{]}, FenceGAN estimates the distance between $p_g(\textbf{x})$
and the tail of $p_{\textbf{x}}(\textbf{x})$. Its limitations are
sampling complexity, the parallel estimation of $p_{\textbf{x}}(\textbf{x})$
and the tail of $p_{\textbf{x}}(\textbf{x})$, and disconnected boundary
generation.

\section{The Proposed TailGAN}

Section~II.B has presented GANs for anomaly detection that perform
sample generation on the support boundary of the normal data distribution.
In this section, we present TailGAN for sample generation on the tail
and anomaly detection.

\subsection{Boundary Generation: Improvement on Leave-One-Out}

Before proceeding to explain TailGAN, it is important to first motivate
the need to perform accurate sample generation on the tail. The leave-one-out
evaluation, which can be restrictive for evaluating anomaly detection
models, uses the disconnected components of the underlying multimodal
distribution. For any annotated dataset with $K+1$ classes, the classification
problem creates clusters and the decision criterion is a boundary
for classification \cite{ADwithGANs-1}. In high-dimensional spaces,
classes form clusters and are disconnected components. Each cluster
can have more than one modes. The disconnected components of the underlying
distribution are usually known during the evaluation of the model
while the modes of the distribution are not. Now, we denote a sample
generated on the distribution's tail by $T(\textbf{z})$, where $\textbf{z} \sim N(\textbf{0}, \textbf{I})$
is the latent space, and using the $l_p$-norm, $||.||_p$, the clustering
algorithm is given by \begin{align}
& \hspace{-2pt} k(T(\textbf{z})) = \text{argmin}_{i = 1, \dots, K} \, \text{dist}(T(\textbf{z}), \textbf{x}_{i}) \label{eq:eqadssssadfsfasEqEqdfs5141} \\
& \hspace{-2pt} \text{where } \text{dist}(T(\textbf{z}), \textbf{x}_{i}) =  \text{min}_{ j = 1, \dots, L} \, || T(\textbf{z}) - \textbf{x}_{i,j} ||_p \\
& \hspace{-2pt} R(T(\textbf{z}), k) = \text{min}_{ j = 1, \dots, L} \, || T(\textbf{z}) - \textbf{x}_{k,j} ||_p \label{eq:eqEqEq5141ddfsg42} \\
& \hspace{-2pt} R(T(\textbf{z}), k) < \text{min}_{i \neq k, i = 1, \dots, K} \, \text{min}_{ j = 1, \dots, L} \, || \textbf{x}_{i,j} - \textbf{x}_{k,j} ||_p \label{eq:eqEqEq5141}
\end{align}where we use $K$ clusters from the leave-one-out methodology, $L$
samples from every class/cluster, and $\textbf{x}_{i,j} \in \mathbb{R}^d$
is the $j$-th sample of class $i$. With our boundary model, we can
decide support membership to improve anomaly detection and also weight
misses and false alarms. We can also generate anomalies, including
adversarial anomalies, and the inequality presented in \eqref{eq:eqEqEq5141}
states that the model's generated samples are closer to the relevant
class than any sample from any other class. Using \eqref{eq:eqEqEq5141ddfsg42}
and \eqref{eq:eqEqEq5141}, our boundary model improves the leave-one-out
evaluation methodology by the margin given by $|R(T(\textbf{z}), k) - \text{min}_{i \neq k, i = 1, \dots, K} \, \text{min}_{ j = 1, \dots, L} \, || \textbf{x}_{i,j} - \textbf{x}_{k,j} ||_p|$.

\subsection{Framework for Sample Generation on the Tail}

In this section, we develop our model, TailGAN, to detect anomalies
near the low probability regions of the data distribution, i.e. strong
anomalies. GANs generally do not guarantee the existence of a probability
density and we use the recently developed PresGAN and FlowGAN models.
We leverage such models for sample generation on the tail using two
steps. The first step is to train either PresGAN or FlowGAN to learn
the ``normal'' distribution, $G(\textbf{z}) \sim p_g(\textbf{x})$.
The random variable $\textbf{z}$ follows a standard Gaussian distribution,
$\textbf{z} \sim N(\textbf{0}, \textbf{I})$, and the mapping from
the latent space, $\textbf{z}$, to the data space, $\textbf{x} \in \mathbb{R}^d$,
is given by $G(\textbf{z})$. The second step is to train a generator,
$T(\textbf{z})$, to perform sample generation on the tail by minimizing
\begin{align}
\begin{split}
L_{tot}(\pmb{\theta}, \textbf{z}, \textbf{x}, G) = & \ w_{pr} \, L_{pr}(\pmb{\theta}, \textbf{z}, G) + w_{d} \, L_d(\pmb{\theta}, \textbf{z}, \textbf{x})\\
& + w_e \, L_e(\pmb{\theta}, \textbf{z}, G) + w_{sc} \, L_{sc}(\pmb{\theta}, \textbf{z})
\end{split}
\label{eq:eqeqeqeq3131}
\end{align}
\label{eq:eqeqeqeq2121}where the total cost function, $L_{tot}$, comprises the probability
cost, $L_{pr}$, the distance loss, $L_{d}$, the maximum-entropy
cost, $L_{e}$, and the scattering loss, $L_{sc}$. The total cost
in \eqref{eq:eqeqeqeq3131} comprises four terms. The probability
cost penalizes probability density to find the tail of the data distribution
while the distance loss penalizes large distance from normality using
the distance from a point to a set. The maximum-entropy loss is for
the dispersion of the samples {[}9,~1{]}, and the scattering cost,
$L_{sc}$, is defined by the ratio of the distances in the $\textbf{z}$
and $\textbf{x}$ spaces to address mode collapse. Hence, $L_{tot}$
in \eqref{eq:eqeqeqeq3131} is given by \begin{align}
\begin{split}
& \ \frac{1}{N} \sum_{i=1}^N \left[ w_{pr} \, p_g( T(\textbf{z}_i; \pmb{\theta}) ) + w_{d} \, \min_{j=1}^M || T(\textbf{z}_i; \pmb{\theta}) - \textbf{x}_j ||_p \right. \\
& \left. + \, w_e \, p_g(T(\textbf{z}_i; \pmb{\theta})) \log(p_g(T(\textbf{z}_i; \pmb{\theta}))) \right. \\
& \left. + \, w_{sc} \, \frac{1}{N-1} \sum_{j=1, \, j \neq i}^N \dfrac{ || \textbf{z}_i - \textbf{z}_j||_p^q }{ || T(\textbf{z}_i; \pmb{\theta}) - T(\textbf{z}_j; \pmb{\theta}) ||_p^q } \right]
\end{split}
\label{eq:eqeq319}
\end{align}
\label{eq:eqeq219}where we leverage the tradeoff between probability and distance in
the first two terms of the loss, i.e. $L_{pr}$ and $L_{d}$, and
where the model probability, $p_g( T(\textbf{z}_i; \pmb{\theta}) )$,
is given by \begin{align}
\label{eq:eqeq219asfgdsffs91}
& \, p_{\textbf{z}}(G^{-1}(T(\textbf{z}_i; \pmb{\theta}))) \left | \text{det} \, \textbf{J}_{G}(T(\textbf{z}_i; \pmb{\theta})) \right |^{-1} \\
= & \exp ( \log (p_{\textbf{z}}(G^{-1}(T(\textbf{z}_i; \pmb{\theta})))) - \log ( \left | \text{det} \, \textbf{J}_{G}(T(\textbf{z}_i; \pmb{\theta})) \right |))\nonumber
\end{align}
\label{eq:eqeq21992}{\setlength{\parindent}{0cm}where $\log (p_{\textbf{z}}(G^{-1}(T(\textbf{z}))))$
and $\log ( \left | \text{det} \, \textbf{J}_{G}(T(\textbf{z})) \right | )$
are estimated by an invertible GAN model such as FlowGAN.}

The parameters of the generator $T(\textbf{z})$, $\pmb{\theta}$,
are obtained by running Gradient Descent on $L_{tot}$, which can
decrease to zero and is written in terms of the sample size, $M$,
and the batch size, $N \leq M$. The distance term of the loss in
\eqref{eq:eqeqeqeq3131}, $L_d(\pmb{\theta}, \textbf{z}, \textbf{x})$,
depends on the training data, $\textbf{x}$. This distance term could
use $G(\textbf{z})$ instead of $\textbf{x}$ and in \eqref{eq:eqeq319},
our distance metric is defined using $M$ by $\text{dist}(T(\textbf{z}_i), \textbf{x}) =  \text{min}_{ j = 1, \dots, M} \, || T(\textbf{z}_i) - \textbf{x}_{j} ||_p$.

The loss in \eqref{eq:eqeq319} uses the $l_p$-norm, $||.||_p$,
and the scattering loss, $L_{sc}$, is based on the $l_p$-norm to
power $q$, $||.||_p^q$, where $p$ and $q$ are real numbers and
$p,q \geq 1$. In \eqref{eq:eqeq319}, the weight $w_{pr}$ is equal
to $1$ and $w_{d}$, $w_{e}$, and $w_{sc}$ are hyperparameters.
In \eqref{eq:eqeqeqeq3131}-\eqref{eq:eqeq219asfgdsffs91}, the gradient
of $L_{tot}$ with respect to $T(\textbf{z})$ is well-defined and
the change of variables formula in \eqref{eq:eqeq219asfgdsffs91}
has been used in a signal processing and nonlinear filtering algorithm
in \cite{key-1}.

\begin{figure}[t]
\vspace{-8.25pt}


\vspace{0.077in}

\centering \includegraphics[clip,width=0.88\columnwidth]{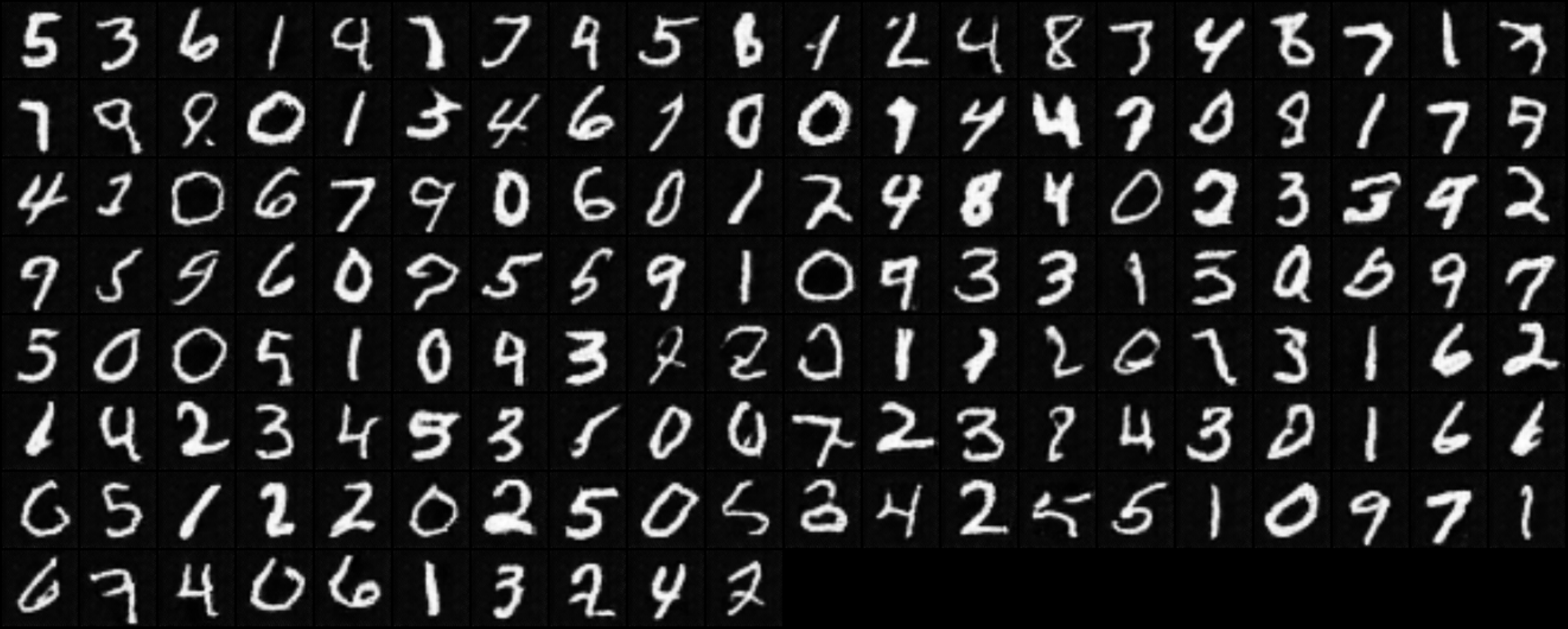}

\vspace{-8pt}

\caption{Generated samples of handwritten digits when we train PresGAN on MNIST
using the PresGAN hyperparameter $\lambda = 0.0002$, at $32$ epochs.}

\label{fig:EDCs-1-2-1-2-1-1-1-2-1-1-1-2-1-1-1-1-1-1-1-1-2-5-001-1-3-1-1-1-1-1-1-1-1-2-1}

\label{fig:EDCs-1-2-1-2-1-1-1-2-1-1-1-2-1-1-1-1-1-1-1-1-2-5-001-1-3-1-1-1-1-1-1-1-1-2-1-1}
\end{figure}

\begin{figure}[t]
\vspace{-8.25pt}

\centering \includegraphics[clip,width=0.61\columnwidth]{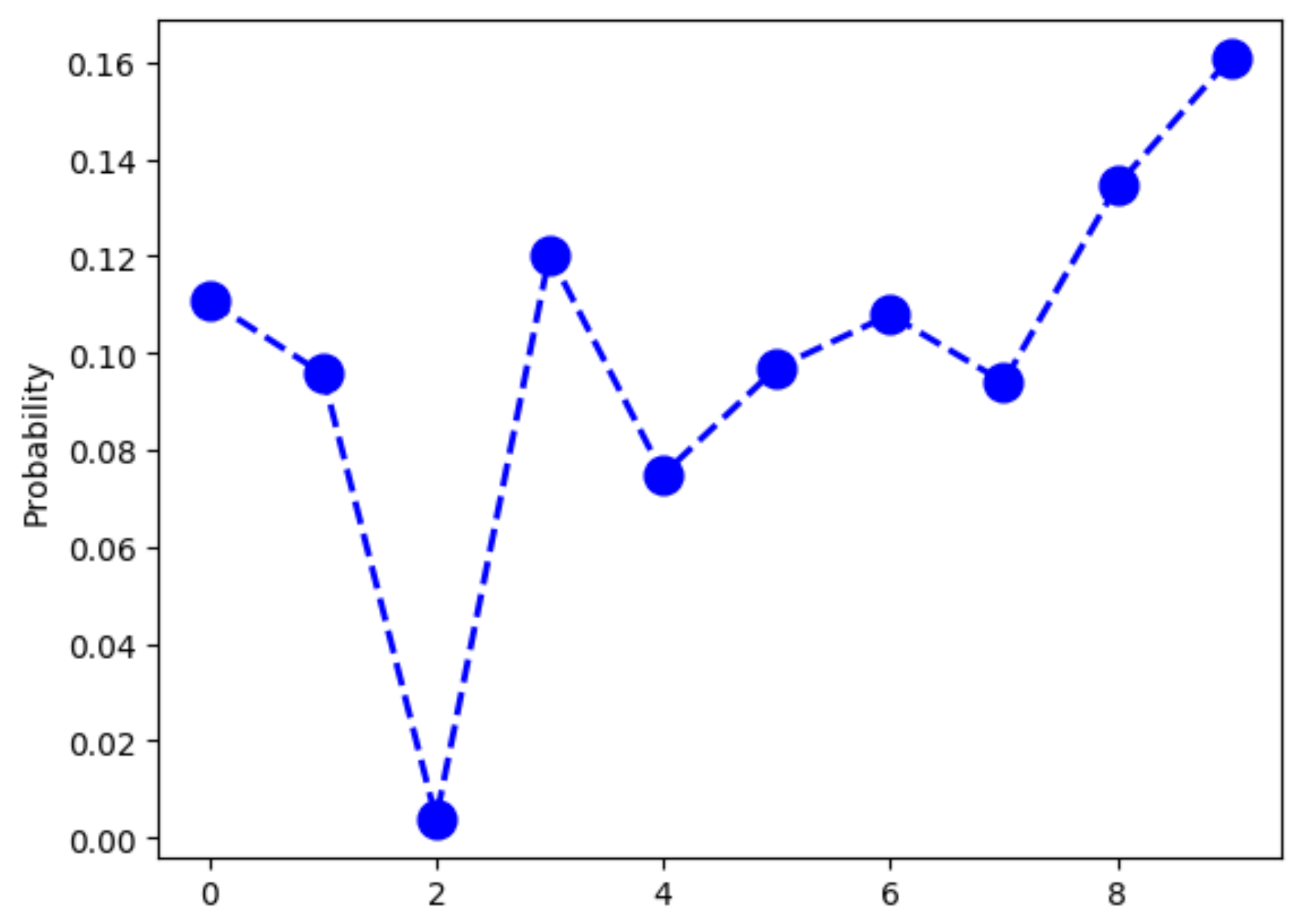}

\vspace{-8pt}

\caption{Probability of occurence of the generated images plotted against the
disconnected component index using (2) and (3) when PresGAN is trained
on MNIST using leave-one-out evaluation. The anomaly class is digit
$2$.}

\label{fig:EDCs-1-2-1-2-1-1-1-2-1-1-1-2-1-1-1-1-1-1-1-1-2-5-001-1-3-1-1-1-1-1-1-1-1-2-1-3-1}

\label{fig:EDCs-1-2-1-2-1-1-1-2-1-1-1-2-1-1-1-1-1-1-1-1-2-5-001-1-3-1-1-1-1-1-1-1-1-2-1-1-2-1}
\end{figure}

\begin{table}[t]
\vspace{-2.25pt}


\vspace{0.077in}

\centering

\begin{tabular}{|c|c|c|c|}
\hline 
{\footnotesize \textbf{MNIST}} & {\footnotesize $L_{tot}$} & {\footnotesize $L_d$} & {\footnotesize $L_{sc}$}\tabularnewline
\hline 
\hline 
{\footnotesize MNIST Digits 0-9} & $4.73$ & $471.96$ & $0.82$\tabularnewline
\hline 
{\footnotesize Fashion-MNIST} & $14.91$ & $1489.70$ & $0.82$\tabularnewline
\hline 
{\footnotesize KMNIST} & $14.78$ & $1476.81$ & $0.82$\tabularnewline
\hline 
\end{tabular}

\vspace{4pt}

\caption{Evaluation of TailGAN comparing normality with anomaly cases using
algorithm convergence criteria, total loss and distance loss. Anomaly
cases: Fashion-MNIST, KMNIST.}

\label{tab:tabTabTab333331sadfsa3sadfs1sadfsafasdf311sdfsa11}

\label{tab:tabTabTab333331sadfsa3sadfs1sadfsafasdf311sdfsa11}
\end{table}

\begin{figure}[t]
\vspace{-8.25pt}

\vspace{-8.25pt}

\vspace{-3.25pt}

\centering \includegraphics[clip,width=0.886\columnwidth]{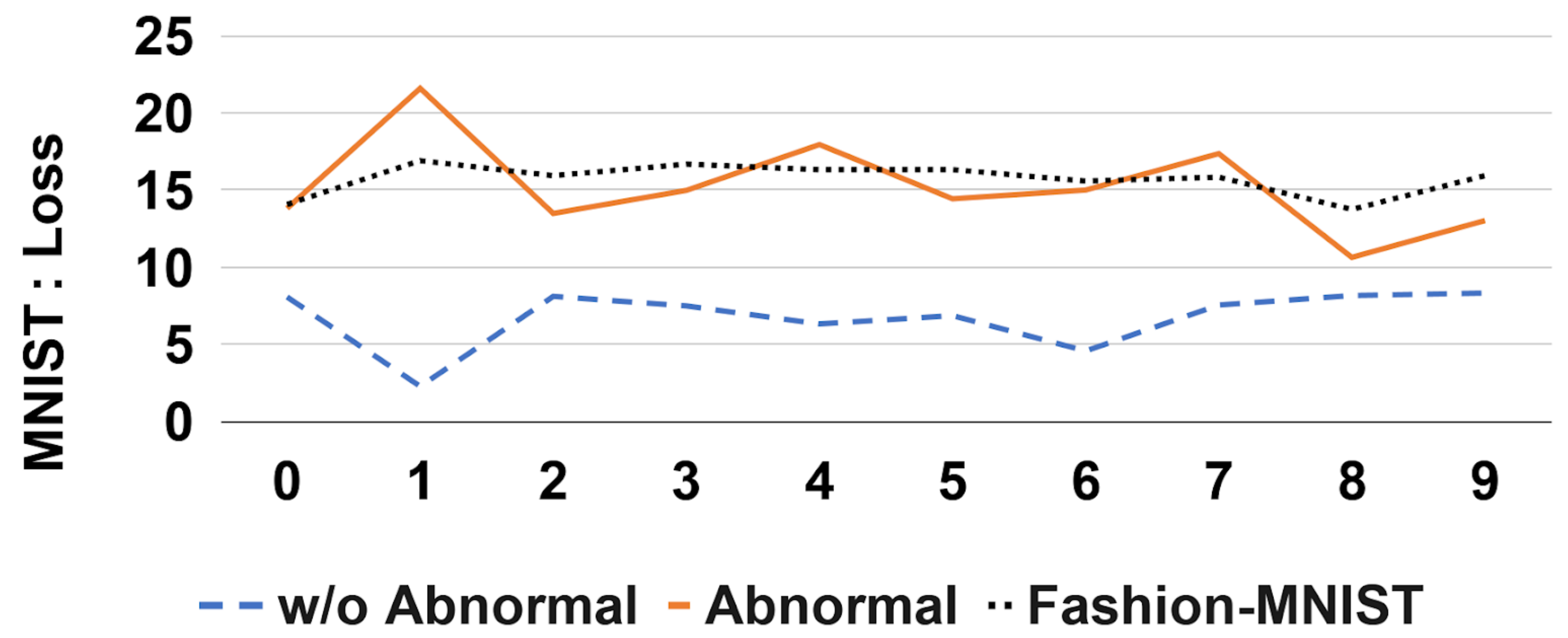}

\vspace{-4pt}

\caption{Leave-one-out evaluation of TailGAN trained on MNIST using $L_{tot}$.
The anomaly cases are the leave-out class and data from Fashion-MNIST.}

\label{fig:EDCs-1-2-1-2-1-1-1-2-1-1-1-2-1-1-1-1-1-1-1-1-2-5-001-1-3-1-1-1-1-1-1-1-1-2-1-4-1-1}

\label{fig:EDCs-1-2-1-2-1-1-1-2-1-1-1-2-1-1-1-1-1-1-1-1-2-5-001-1-3-1-1-1-1-1-1-1-1-2-1-1-3-1-1}
\end{figure}

\begin{figure}[t]
\vspace{-8.25pt}

\vspace{-8.25pt}

\centering \includegraphics[clip,width=1\columnwidth]{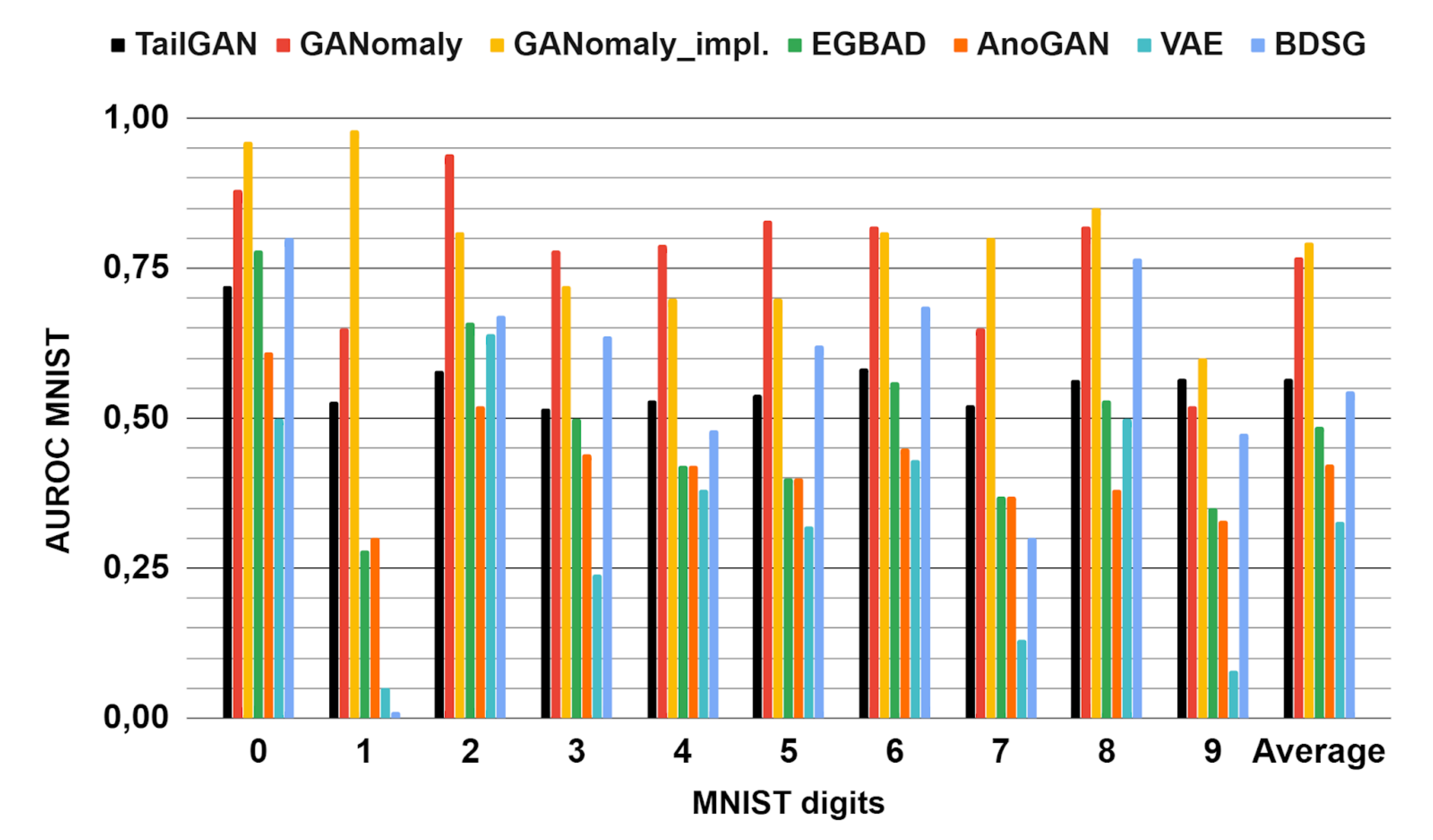}

\vspace{-8pt}

\caption{Evaluation of TailGAN using AUROC on MNIST data.}

\label{fig:EDCs-1-2-1-2-1-1-1-2-1-1-1-2-1-1-1-1-1-1-1-1-2-5-001-1-3-1-1-1-1-1-1-1-1-2-1-2-1-2}

\label{fig:EDCs-1-2-1-2-1-1-1-2-1-1-1-2-1-1-1-1-1-1-1-1-2-5-001-1-3-1-1-1-1-1-1-1-1-2-1-1-1-1-2}
\end{figure}

\begin{figure}[t]
\vspace{-8.25pt}

\centering \includegraphics[clip,width=1\columnwidth]{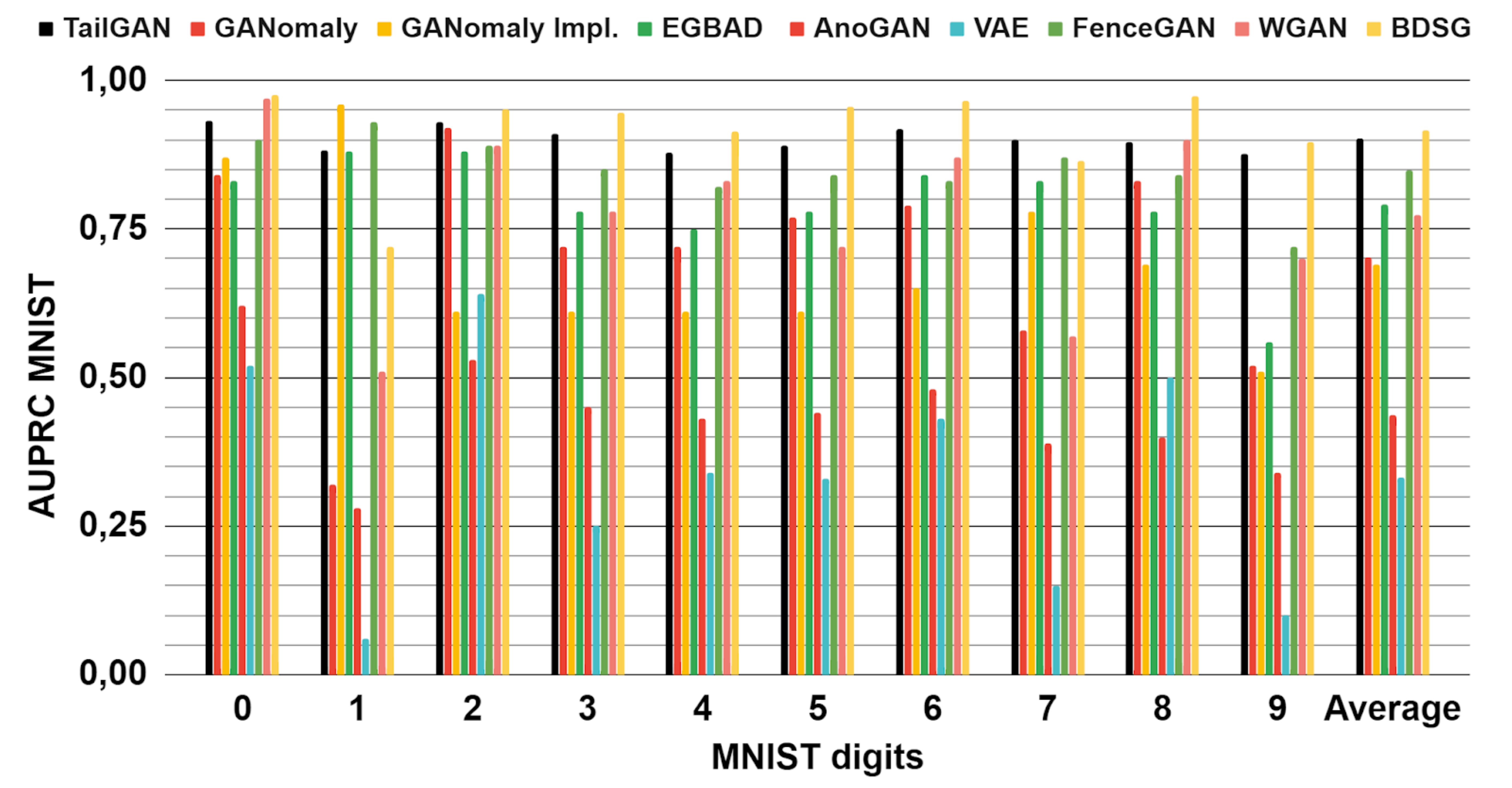}

\vspace{-8pt}

\caption{Evaluation using AUPRC for anomaly detection on MNIST.}

\label{fig:EDCs-1-2-1-2-1-1-1-2-1-1-1-2-1-1-1-1-1-1-1-1-2-5-001-1-3-1-1-1-1-1-1-1-1-2-1-2-1-1-1}

\label{fig:EDCs-1-2-1-2-1-1-1-2-1-1-1-2-1-1-1-1-1-1-1-1-2-5-001-1-3-1-1-1-1-1-1-1-1-2-1-1-1-1-1-1}
\end{figure}

\begin{figure}[t]
\vspace{-8.25pt}

\vspace{-1.25pt}

\centering \includegraphics[clip,width=1\columnwidth]{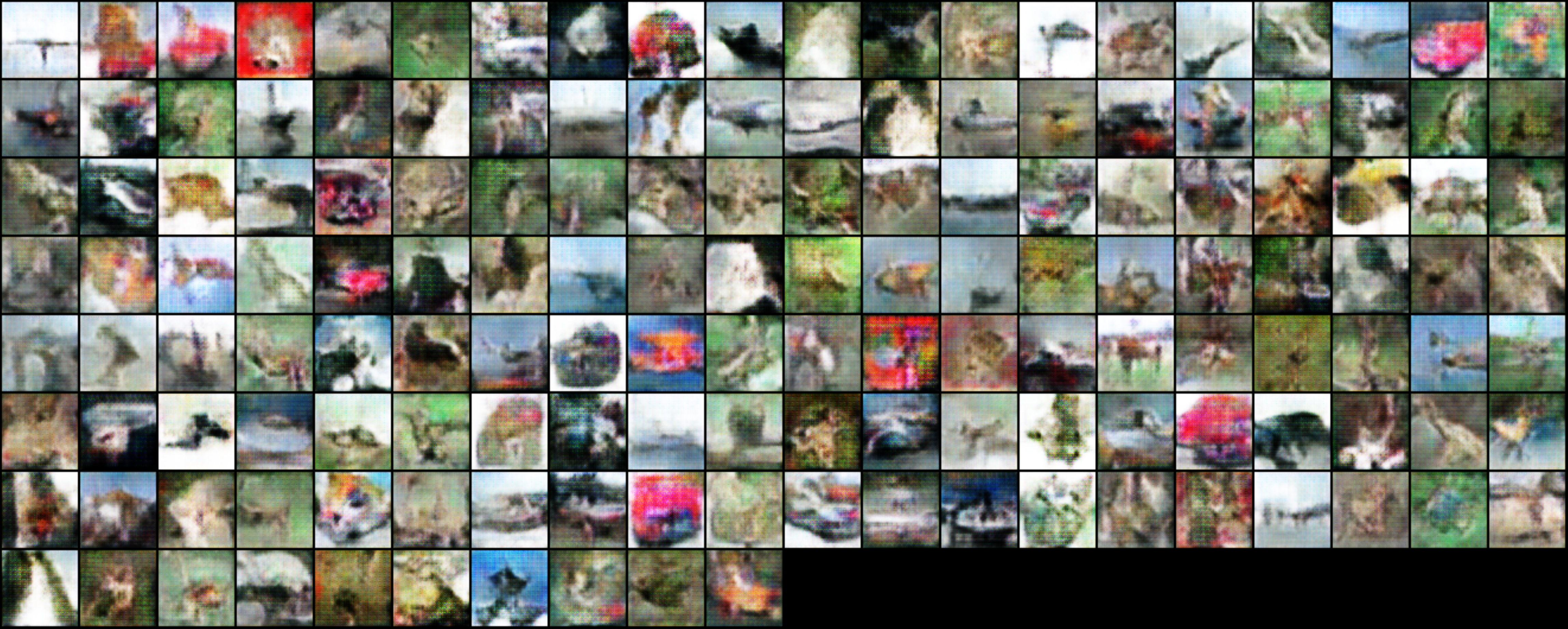}

\vspace{-8pt}

\caption{Generated samples when we train PresGAN on CIFAR-10 data.}

\label{fig:EDCs-1-2-1-2-1-1-1-2-1-1-1-2-1-1-1-1-1-1-1-1-2-5-001-1-3-1-1-1-1-1-1-1-1-2-1-2}

\label{fig:EDCs-1-2-1-2-1-1-1-2-1-1-1-2-1-1-1-1-1-1-1-1-2-5-001-1-3-1-1-1-1-1-1-1-1-2-1-1-1}
\end{figure}

\section{Evaluation of TailGAN}

We evaluate the TailGAN model using (i) algorithm convergence criteria
such as the value of the objective cost function in \eqref{eq:eqeq319}
and the value of the distance cost, $L_{d}$, which is the distance
from a point to a set and is based on the $l_p$-norm and the minimum
operator, and (ii) the Area Under the Receiver Operating Characteristics
Curve (AUROC) and the Area Under the Precision-Recall Curve (AUPRC).
Experiments are performed on datasets of increasing complexity, MNIST,
CIFAR, and Baggage X-Ray. In contrast to \cite{Deecke}, we use the
leave-one-out evaluation methodology and the detection of abnormal
OoD data. The leave-one-out evaluation that uses the leave-out class
as the anomaly leads to multimodal distributions with a support with
disjoint components. The boundary of the support of the data distribution
is defined using the threshold $\epsilon$.

TailGAN performs efficient sample generation on the tail of the distribution
obviating the rarity sampling complexity problem, not requiring importance
sampling \cite{ADwithGANs-1}. For distributions with disconnected
components, the TailGAN model achieves better performance than the
convex hull extrema points.

\subsection{Implementation of the Proposed TailGAN}

We implement TailGAN in PyTorch and a vectorized implementation of
the individual terms of the cost function in \eqref{eq:eqeq319} has
been created. We evaluate TailGAN using $p=q=2$ and because the choice
of distance metric is important, $p=q=1$ or $p=2$ and $q=1$ could
be used. We use $p_g(\textbf{x})$ from PresGAN which computes the
entropy to address mode collapse. The first connection between TailGAN
and our chosen base mode is $p_g(\textbf{x})$ while the second connection
is model initialization, $\pmb{\theta}_{t0} = \pmb{\theta}_{g}$,
where $\pmb{\theta}_{t0}$ are the parameters of the generator $T(\textbf{z})$
at the start of training and $\pmb{\theta}_g$ are the parameters
of $G(\textbf{z})$.

Using $\pmb{\theta}_{t0} = \pmb{\theta}_{g}$, $T(\textbf{z})$ is
trained to perform sample generation on the tail of the normal data
distribution by starting from within the distribution, and this differs
from the encoder-based initialization used in \cite{Deecke}. On the
contrary, using random initialization for $\pmb{\theta}$, $T(\textbf{z})$
is trained to perform sample generation on the tail by starting from
outside the distribution. To compute the probability density, $p_g(\textbf{x})$,
in \eqref{eq:eqeq319} from the entropy which is estimated by PresGAN,
we use the Lambert $W(.)$ function and Newton\textquoteright s iterations.
We evaluate TailGAN by first performing density estimation and then
training $T(\textbf{z})$ using convolutional layers with batch normalization,
minimizing \eqref{eq:eqeqeqeq3131} and  \eqref{eq:eqeq319}.

\subsection{Evaluation of TailGAN on MNIST Data}

We first train PresGAN on MNIST until convergence using the leave-one-out
methodology and the detection of abnormal OoD data. Next, we train
TailGAN applying the objective cost function in \eqref{eq:eqeq319}.
We examine different values for the batch size, $N$, and we use the
entire training set for the sample size, $M$. We create $T(\textbf{z})$
using convolutional networks and we examine different architectures
such as feed-forward and residual.

Figure~1 shows the generated samples at $32$ epochs when we train
and use a modified version of PresGAN and the PresGAN hyperparameter
$\lambda = 0.0002$. As qualitative evaluation and visual measure,
the MNIST canvas with the generated images in Fig.~1 shows that our
chosen base model is trained to create realistic images of handwritten
digits. All the digits from $0$ to $9$ are present in the canvas.
Figure~2 depicts the probability of occurence of the generated images
against the disconnected component index when PresGAN is trained on
MNIST using the leave-one-out evaluation, when the anomaly class is
digit $2$. The frequency of the generated data is computed using
(2) and (3) based on the clustering algorithm in Sec.~III.A.

We train TailGAN on MNIST until convergence and all the individual
terms of the objective cost function, i.e. $L_{pr}$, $L_{d}$, $L_{e}$,
and $L_{sc}$, decrease over epochs achieving convergence.

For the evaluation of TailGAN, Table~I shows the algorithm convergence
criteria in \eqref{eq:eqeqeqeq3131}-\eqref{eq:eqeq219asfgdsffs91}
produced by TailGAN trained on MNIST data. The values of the objective
cost function, total loss, and distance loss, $L_{d}$, for abnormal
OoD data are higher than the corresponding values for the normal data.
The OoD anomalies are from Fashion-MNIST and KMNIST, and the total
and distance losses are indicators of anomalies.

Using the leave-one-out evaluation on MNIST, we compute the cost function
in (7), the total cost, the distance loss, and the scattering cost
of $T(\textbf{z})$ of TailGAN. Figure~3 depicts the total loss,
$L_{tot}$, for every MNIST digit for three different cases: (i) Without
the abnormal digit, (ii) The abnormal digit, and (iii) Fashion-MNIST.
The anomaly cases are the abnormal leave-out digit and Fashion-MNIST
data. The algorithm convergence criteria, $L_{tot}$ and $L_{d}$,
deviate from normality for the abnormal digits and the OoD cases as
they are higher compared to the corresponding normal case values.
Comparing TailGAN to the GANomaly and FenceGAN baselines trained on
MNIST and evaluated on Fashion-MNIST, the $l_2$-norm distance loss
in the $\textbf{x}$ space is $3.1$, $1.7$, and $1.1$ times as
much compared to the corresponding values for the normal case for
TailGAN (i.e. Table~I), GANomaly, and FenceGAN, respectively.

Figures~4 and 5 present the AUROC and AUPRC scores, respectively,
for the evaluation of TailGAN on MNIST. During inference, we use the
estimated probability density, the first term in \eqref{eq:eqeq319}
which addresses within or out of the distribution support, and the
second term in \eqref{eq:eqeq319} which computes $l_p$-norm distances
for the anomaly score. Figures~4 and 5 examine the performance of
TailGAN using the leave-one-out evaluation compared to several baselines,
GANomaly, the implementation of GANomaly (GANomaly Impl.), EGBAD,
AnoGAN, VAE, FenceGAN, and BDSG which has been proposed in \cite{AdamOptimizer-1}.

\begin{table}[t]
\centering

\begin{tabular}{|c|c|c|c|}
\hline 
{\footnotesize \textbf{CIFAR-10}} & {\footnotesize $L_{tot}$} & {\footnotesize $L_d$} & {\footnotesize $L_{sc}$}\tabularnewline
\hline 
\hline 
{\footnotesize CIFAR-10} & $2.22$ & $217.72$ & $4.60$\tabularnewline
\hline 
{\footnotesize CIFAR-100} & $5.21$ & $516.49$ & $4.60$\tabularnewline
\hline 
{\footnotesize SVHN} & $7.76$ & $771.02$ & $4.60$\tabularnewline
\hline 
{\footnotesize STL-10} & $5.74$ & $569.21$ & $4.60$\tabularnewline
\hline 
{\footnotesize CelebA} & $6.31$ & $626.26$ & $4.60$\tabularnewline
\hline 
{\footnotesize Baggage X-Ray} & $51.67$ & $5161.90$ & $4.60$\tabularnewline
\hline 
\end{tabular}

\vspace{4pt}

\caption{Evaluation of TailGAN comparing normal with abnormal cases using algorithm
convergence criteria, total loss and distance loss. Normality: CIFAR-10
classes 0-9. Anomaly cases: CIFAR-100, SVHN, STL-10, CelebA, Baggage
X-Ray. }

\label{tab:tabTabTab333331sadfsa3sadfs1sadfsafasdf311sdfsa11}

\label{tab:tabTabTab333331sadfsa3sadfs1sadfsafasdf311sdfsa11}
\end{table}

\begin{figure}[t]
\vspace{-8.25pt}

\begin{minipage}[t]{0.48\columnwidth}%
\centering \includegraphics[clip,width=0.73\columnwidth]{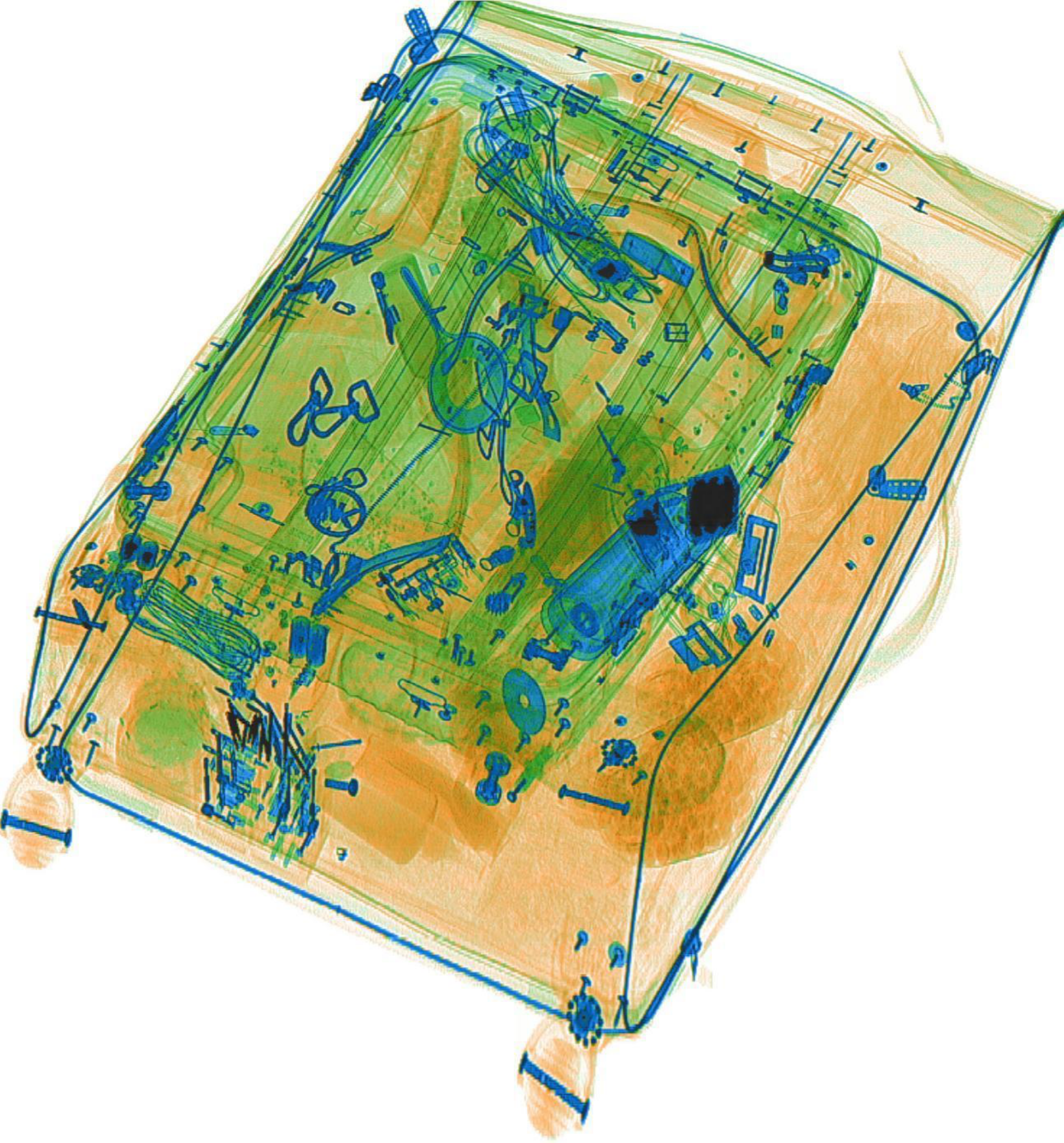}

\vspace{-2pt}

{\footnotesize (a)}%
\end{minipage}\hfill{}%
\begin{minipage}[t]{0.48\columnwidth}%
\centering \includegraphics[clip,width=1\columnwidth]{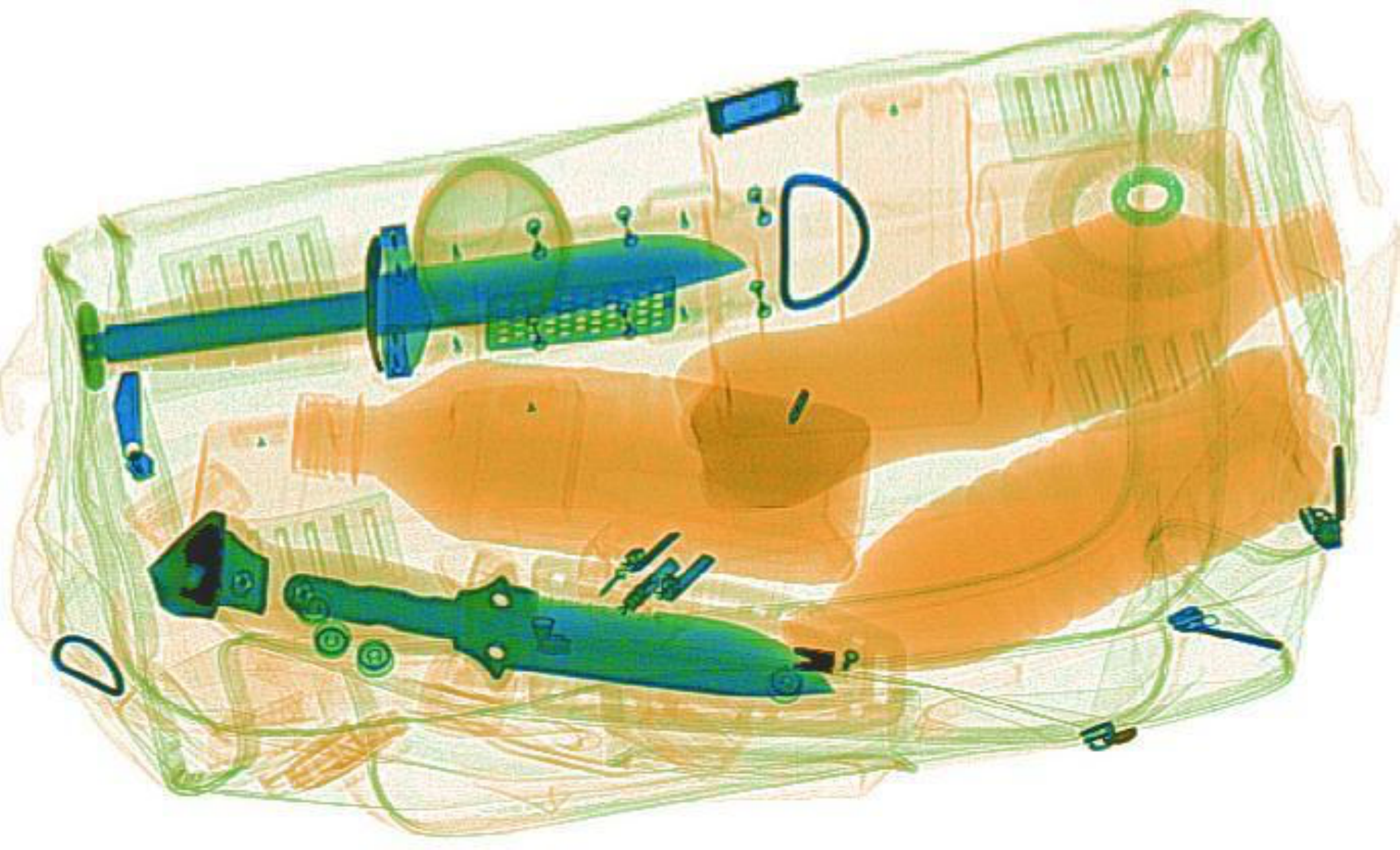}

\vspace{-2pt}

{\footnotesize (b)}%
\end{minipage}

\vspace{-2pt}

\caption{(a) Normal class image from the Baggage X-Ray dataset. (b) Anomaly
class image from the Baggage X-Ray dataset containing knives and bottles.}

\label{fig:EDCs-1-2-1-2-1-1-1-2-1-1-1-2-1-1-1-1-1-1-1-1-2-5-001-1-3-1-1-1-1-1-1}

\label{fig:EDCs-1-2-1-2-1-1-1-2-1-1-1-2-1-1-1-1-1-1-1-1-2-5-001-1-3-1-1-1-1-1-1-1}

\label{fig:EDCs-1-2-1-2-1-1-1-2-1-1-1-2-1-1-1-1-1-1-1-1-2-5-001-1-3-1-1-1-1-1-1-1-2}

\label{fig:EDCs-1-2-1-2-1-1-1-2-1-1-1-2-1-1-1-1-1-1-1-1-2-5-001-1-3-1-1-1-1-1-1-1-1}

\label{fig:EDCs-1-2-1-2-1-1-1-2-1-1-1-2-1-1-1-1-1-1-1-1-2-5-001-1-3-1-1-1-1-1-1-1-1-1}

\label{fig:EDCs-1-2-1-2-1-1-1-2-1-1-1-2-1-1-1-1-1-1-1-1-2-5-001-1-3-1-1-1-1-1-1-1-1-1-1}
\end{figure}

\begin{table}[t]
\centering

\begin{tabular}{|c|c|c|c|}
\hline 
{\footnotesize \textbf{Baggage X-Ray}} & {\footnotesize $L_{tot}$} & {\footnotesize $L_d$} & {\footnotesize $L_{sc}$}\tabularnewline
\hline 
\hline 
{\footnotesize Baggage X-Ray Normal} & $3.21$ & $319.22$ & $2.83$\tabularnewline
\hline 
{\footnotesize Baggage X-Ray Abnormal} & $7.45$ & $741.39$ & $2.83$\tabularnewline
\hline 
{\footnotesize CIFAR-10} & $12.77$ & $1274.32$ & $2.83$\tabularnewline
\hline 
{\footnotesize CIFAR-100} & $13.14$ & $1309.87$ & $2.83$\tabularnewline
\hline 
{\footnotesize SVHN} & $9.87$ & $983.72$ & $2.83$\tabularnewline
\hline 
{\footnotesize STL-10} & $17.11$ & $1706.81$ & $2.83$\tabularnewline
\hline 
\end{tabular}

\vspace{4pt}

\caption{Evaluation of TailGAN on Baggage X-Ray data comparing normality with
abnormal OoD cases using $L_{tot}$ from (7). }

\label{tab:tabTabTab333331sadfsa3sadfs1sadfsafasdf311sdfsa11}

\label{tab:tabTabTab333331sadfsa3sadfs1sadfsafasdf311sdfsa11}
\end{table}

\subsection{Evaluation of TailGAN on CIFAR-10 Data}

To scale up the dimensions of the problem, we use the CIFAR-10 dataset
and we hence go from $28 \times 28$ dimensions of MNIST to $3 \times 32 \times 32$
dimensions of CIFAR. First, we train PresGAN for density estimation
until convergence on CIFAR data and, then, we train TailGAN using
convolutional neural networks with batch normalization. We train TailGAN
using the entire training set for $M$ and our aim is to use TailGAN
to accurately detect atypical, aberrant, abnormal samples.

Figure~6 shows the generated samples when we use a modified version
of PresGAN and the PresGAN hyperparameter $\lambda = 0.0002$. As
qualitative evaluation, the CIFAR-10 canvas with the generated images
in Fig.~6 shows that our chosen base model is trained to create realistic
images. All the classes are present in the canvas. Next, we train
the proposed TailGAN by minimizing the cost function presented in
\eqref{eq:eqeqeqeq3131}-\eqref{eq:eqeq219asfgdsffs91}. TailGAN achieves
convergence on CIFAR-10 and $L_{pr}$, $L_{d}$, $L_{e}$, and $L_{sc}$
in \eqref{eq:eqeqeqeq3131} and \eqref{eq:eqeq319} decrease over
iterations, epochs, and time.

For model evaluation during inference, Table~II compares the values
of the loss terms for the normal class, CIFAR-10 data, with the corresponding
values of the loss terms for the abnormal OoD data, CIFAR-100, SVHN,
STL-10, CelebA, and Baggage X-Ray. In Table~II, the distance loss
is an indicator of anomalies and of the anomaly score, and both the
total and distance losses are indicators of anomalies. The distance
loss and the cost function values, $L_{d}$ and $L_{tot}$, are higher
for the OoD data than the corresponding values for normality: $L_{tot}=2.22$
for normal data from CIFAR-10 and $L_{tot}=6.31$ for abnormal data
from CelebA. Comparing TailGAN with the GANomaly and FenceGAN baselines
trained on CIFAR-10 (Normal) and evaluated on CIFAR-100 (Abnormal),
the $l_2$-norm distance loss in the $\textbf{x}$ space is $2.3$,
$1.1$, and $1.1$ times as much compared to the corresponding values
for normality for TailGAN (i.e. Table~II), GANomaly, and FenceGAN.

\subsection{Evaluation of TailGAN on Baggage X-Ray Data}

In this section, we evaluate TailGAN on Baggage X-Ray data. Figure~7
shows two example images, one from the normal class and one from the
anomaly class, from the Baggage X-Ray dataset, where the abnormal
image contains $2$ knives and $3$ bottles. Table~III uses Baggage
X-Ray data and compares the values of the losses for the normal class
with the values of the losses for the abnormal OoD data. The distance
loss and the cost function values, $L_{d}$ and $L_{tot}$, are higher
for the Baggage X-Ray abnormal OoD data than the corresponding values
for the normal Baggage X-Ray data, i.e. $L_{tot}=3.21$ for normal
class data and $L_{tot}=7.45$ for abnormal OoD image data.

\section{Conclusion}

In this paper, we have proposed the TailGAN model to perform anomaly
detection and sample generation on the tail of the distribution of
typical samples. The proposed TailGAN uses adversarial training and
GANs, and minimizes the objective cost function in \eqref{eq:eqeqeqeq3131}-\eqref{eq:eqeq219asfgdsffs91}.
Using GANs that can explicitly compute the probability density, we
perform sample generation on the tail of the data distribution, we
address multimodal distributions with disconnected-components supports,
and we also address the mode collapse problem. The main evaluation
outcomes on MNIST, CIFAR-10, Baggage X-Ray, and OoD data using the
leave-one-out evaluation show that TailGAN achieves competitive anomaly
detection performance.

\section{Acknowledgment}

This work was supported by the Engineering and Physical Sciences Research
Council of the UK (EPSRC) Grant number EP/S000631/1 and the UK MOD
University Defence Research Collaboration (UDRC) in Signal Processing.

\end{document}